\documentclass[sigconf]{acmart}
\copyrightyear{2024}
\acmYear{2024}
\setcopyright{acmlicensed}
\acmConference[WSDM '24] {Proceedings of the 17th ACM International Conference on Web Search and Data Mining}{March 4--8, 2024}{Merida, Mexico.}
\acmBooktitle{Proceedings of the 17th ACM International Conference on Web Search and Data Mining (WSDM '24), March 4--8, 2024, Merida, Mexico}
\acmPrice{15.00}
\acmISBN{979-8-4007-0371-3/24/03}
\acmDOI{10.1145/3616855.3635771}
\usepackage{colortbl}
\definecolor{mygray}{gray}{0.6}
\usepackage{multirow}
\usepackage{latexsym}
\usepackage{subfig}
\usepackage{graphicx}
\usepackage{algorithm}
\usepackage{algorithmic}
\usepackage{tikz}
\usepackage[most]{tcolorbox}
\usepackage{booktabs} 
\usepackage{amsmath,amsfonts}
\usepackage[export]{adjustbox}
\usepackage{tcolorbox}
\usepackage{enumitem}
\setlist{leftmargin=2mm}
\usepackage{pifont}
\usepackage{setspace}
\usepackage{multirow}

\usepackage[normalem]{ulem}
\useunder{\uline}{\ul}{}

\tcbset{
  before skip=0.5\baselineskip,
  after skip=0.5\baselineskip
}

\usepackage[draft,textsize=footnotesize,textwidth=15mm]{todonotes}
\usepackage{balance}

\usepackage[]{geometry}

\begin{document}
\usetikzlibrary{arrows.meta}
\title{Causality Guided Disentanglement for Cross-Platform \\ 
Hate Speech Detection}

\author{Paras Sheth\textsuperscript{\rm 1}, Raha Moraffah\textsuperscript{\rm 1}, Tharindu Kumarage\textsuperscript{\rm 1}, Aman Chadha\textsuperscript{\rm 2}, Huan 
liu\textsuperscript{\rm 1}}
\affiliation{ \textsuperscript{\rm 1}\institution{Computer Science and Engineering, Arizona State University}\state{Arizona}\country{USA} \\
 \textsuperscript{\rm 2}\institution{Amazon Alexa AI}\city{Sunnyvale}\country{USA}}
\email{{psheth5,rmoraffa,kskumara,huanliu}@asu.edu, hi@aman.ai}

\begin{abstract}
Despite their value in promoting open discourse, social media platforms are often exploited to spread harmful content. Current deep learning and natural language processing models used for detecting this harmful content rely on domain-specific terms affecting their ability to adapt to generalizable hate speech detection. This is because they tend to focus too narrowly on particular linguistic signals or the use of certain categories of words. Another significant challenge arises when platforms lack high-quality annotated data for training, leading to a need for cross-platform models that can adapt to different distribution shifts. Our research introduces a cross-platform hate speech detection model capable of being trained on one platform’s data and generalizing to multiple unseen
platforms. One way to achieve good generalizability across platforms is to disentangle the input representations into invariant and platform-dependent features. We also argue that learning causal relationships, which remain constant across diverse environments, can significantly aid in understanding invariant representations in hate speech. By disentangling input into platform-dependent fea-
tures (useful for predicting hate targets) and platform-independent features (used to predict the presence of hate), we learn invariant representations resistant to distribution shifts. These features are then used to predict hate speech across unseen platforms. Our extensive experiments across four platforms highlight our model’s enhanced efficacy compared to existing state-of-the-art methods in
detecting generalized hate speech.

\end{abstract}

\maketitle

\section{Introduction}
\textbf{Warning:} \textit{this paper contains contents that may be upsetting.}

The worldwide adoption of social media platforms has led to a proliferating volume of social exchanges. The communications happening over these platforms can strongly influence public opinions. One form of communication influencing social beliefs and opinions is hate speech. As the name suggests, hate speech refers to those communications that encourage or support acts of violence, prejudice, hostility, or discrimination against an individual or a group based on social characteristics such as race, ethnicity, religion, etc. Hate speech content uses derogatory and insulting language to denigrate specific people or groups. The effects of hate speech extend beyond mental distress for its victims. Numerous studies have demonstrated the link between hate speech, real-world violence, and hate crimes~\cite{akca2020planting,waseem2016hateful}. Thus, given the heinous effects and the aftermath of hate speech, it becomes imperative to detect such content on online social media platforms and prevent its spread for a respectful and sustainable online environment.

\begin{figure}[]
     \centering
     \includegraphics[width=0.35\textwidth]{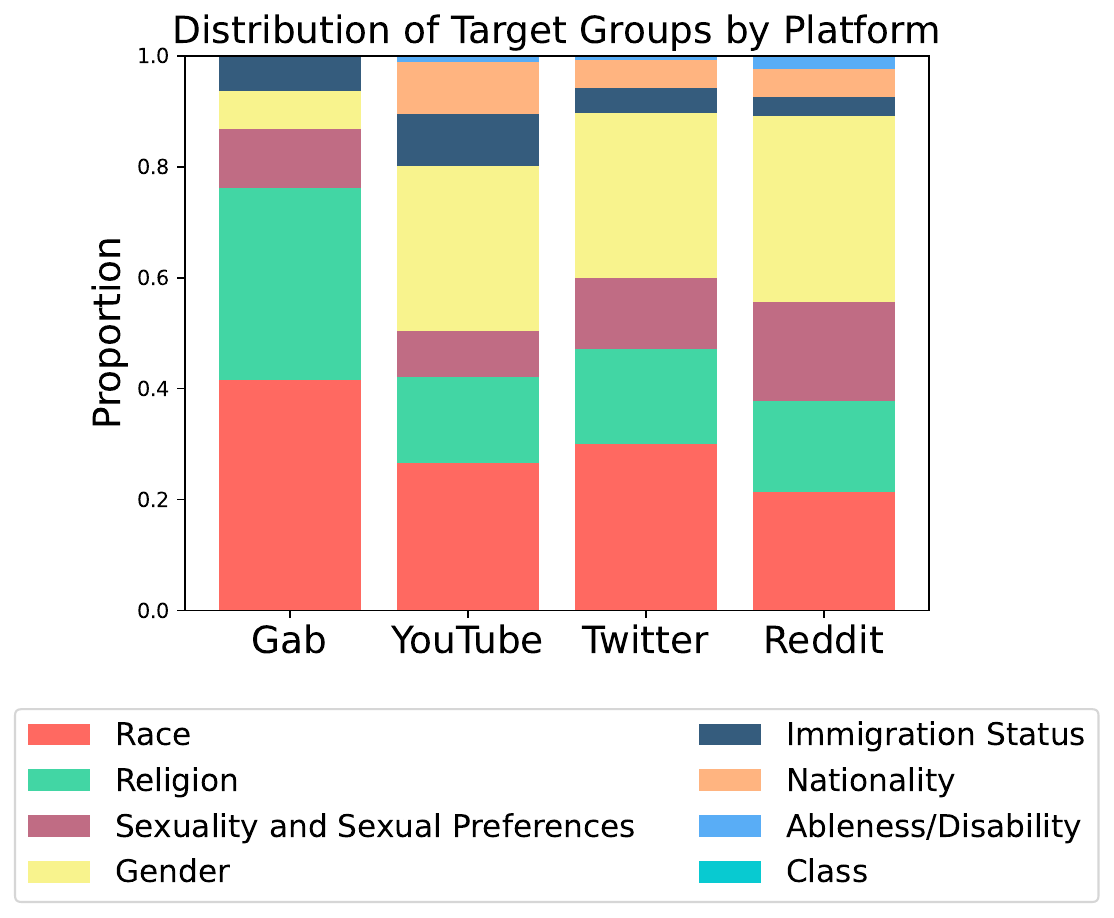}
     \caption{The distribution of the target of hate speech varies across different platforms, indicating that the target can be leveraged as a platform-dependent feature.}
     \vspace{1mm}
\label{fig:intro}
\end{figure}
Due to the volume of online content, it is infeasible to detect hate speech manually. Thus, researchers leverage deep learning and natural language processing-based models to automate hate speech detection~\cite{jahan2023systematic}. However, even automatic hate speech detection is hindered by several existing challenges. One major challenge is the \textit{limited availability of labeled data} for some platforms. Over the recent years, there has been a frequent emergence of new social media platforms~\cite{jeong2023exploring}. Not all these platforms have access to quality annotated data. As a result, to employ hate speech detection models for these platforms, one sensible approach is to leverage available quality training data from other platforms (source). However, if the model does not adjust for the biases present in the training data of the source platform, the model may end up failing to detect hate speech on the target platform~\cite{malik2022deep}. Another critical challenge that hinders the performance of the existing models is the \textit{over-reliance} on linguistic cues and certain category of words (e.g., profane words) to detect hate speech. A range of work~\cite{ramponi-tonelli-2022-features,zhou2021challenges} show that the current state-of-the-art models overly rely on identity words (e.g., "jews") to predict hateful content. Moreover, some studies~\cite{fortuna2020toxic,chiu2021detecting} highlight how the existing models consider content with profane words to be mostly hateful. The replication of such biases can seriously impact the model's capabilities to perform well across platforms. Thus, there is a need for addressing these challenges to enable effective cross-platform generalization for hate speech detection.


A cross-platform hate speech detection model is trained on data from one source domain or platform and applied  to unseen target domains. These cross-platform  models seek to identify the underlying patterns in hate speech rather than flagging specific words or phrases often associated with the content on a platform. This approach is advantageous as it adapts to changes in how different people use language, avoiding over-reliance on specific words~\cite{ramponi-tonelli-2022-features} and aids social media platforms that lack quality labeled data. 

Recent works aim to build such generalizable models for detecting hate. For instance, the authors of~\cite{kim2022generalizable} leverage auxiliary information and contrastive learning to determine whether a post is implicitly hateful. However, the auxiliary information may not be available for large datasets or different platforms, limiting the framework's applicability. Another line of work~\cite{markov2021exploring,min2023finding} considered the POS tags and the emotions for training a generalizable hate speech model. Such methods tend to form a spurious correlation with certain tags (e.g., adverbs) or certain kinds of words (e.g., profane words), thus degrading its generalization power. The authors of~\cite{sheth2023peace} identified two quantifiable invariant cues, namely, aggression and overall sentiment present in text to guide representations to be generalizable for hate speech detection. 
This approach identifies cues manually, however,  more cues might exist that can help guide the representations.
Pretrained language models such as BERT have also been fine-tuned on large hate speech datasets to create specialized hate detection models, such as HateBERT~\cite{caselli2020hatebert} and HateXplain~\cite{mathew2021hatexplain}. However, these models tend to replicate any biases present in the datasets they were fine-tuned on, which can limit their effectiveness when applied to new or different contexts.

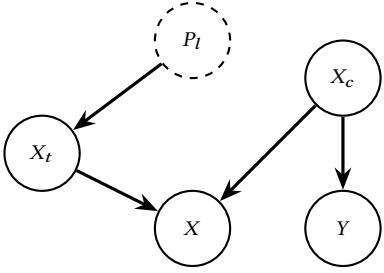
\begin{figure}[]
     \centering
     \footnotesize
     \small
    \begin{tikzpicture}
    \begin{scope}[every node/.style={circle,thick,draw,minimum size=1cm}]
        \node (X) at (-2,0.5) {$X$};
        \node [dashed] (P) at (-2,3) {$P_{l}$};
        \node (Xt) at (-4,1.5) {$X_{t}$};
        \node (Xc) at (0,2.5) {$X_{c}$};
        \node (Y) at (0,0.5) {$Y$};
    \end{scope}
    
    \begin{scope}[>={Stealth[black]},
                  every edge/.style={draw=black,very thick}]
        \path [->] (P) edge node {} (Xt);
        \path [->] (Xt) edge node {} (X);
        \path [->] (Xc) edge node {} (X);
        \path [->] (Xc) edge node {} (Y);
    \end{scope}
    \end{tikzpicture}
     \caption{The causal graph representing the data generating mechanism for hate speech detection. $X_{c}$ represents the causal representation useful for predicting whether the content is hateful or not, $Y$ represents the hate label, $X_{t}$ the target of hate speech, $X$ the input, and $P_{l}$ the latent platform variable influencing the target.}
\label{causal_graph}
\end{figure}
In light of the intricate nature of hate speech detection and the challenges posed by existing state-of-the-art models that strive for cross-platform hate speech detection, we propose to explore ageneralizable hate speech detection model that overcomes these limitations. A prevalent and proven technique for constructing a generalizable model involves leveraging the concept of disentanglement. This paper is the first attempt to adapt the disentanglement technique to cross-platform hate speech detection. Disentanglement segregates the input representation into a platform-dependent component containing platform-specific information and an invariant component containing information shared across different platforms. Causality is recently explored to capture the invariance across different platforms~\cite{sheth2022domain,mahajan2021domain}. Causal features represent the causal relationships or mechanisms underlying a particular phenomenon. When inferred that a variable causes another variable, it is implied that a universal and consistent relationship exists. Identifying and separating such factors allows us to understand the true underlying structure of the data. Furthermore, causal relations are known to be generally invariant, thus can enhance the generalization capabilities of machine learning models~\cite{buhlmann2020invariance}. Thus, a causality-guided disentanglement model demands identifying the platform-dependent features w.r.t. hate speech detection problem.

Investigating the recent empirical analysis~\cite{bourgeade2023did} brings to attention that the distribution of the targets of hate speech varies across platforms, allowing the target of hate speech to be a potential candidate for platform-dependent features. To verify this finding, we considered hate speech datasets from four platforms, i.e., Gab, YouTube, Twitter, and Reddit. The distribution of the target categories across platforms can be seen in Figure~\ref{fig:intro}. It is evident that a platform can influence the choice of the target of hate speech. This is aligned with previous findings where researchers showed platforms such as YouTube, and Reddit are more susceptible to hate speech based on people's gender~\cite{doring2019male,farrell2019exploring} and platforms such as Twitter and GAB are more susceptible to hate speech based on race~\cite{waseem2016you,mathew2019spread}. Thus, considering the target as a platform-dependent variable can facilitate the disentanglement process. To this end, we propose to formulate the data-generating mechanism of the hate speech problem using the causal graph as shown in Figure~\ref{causal_graph}. The causal graph aligns with the intuition that the core properties of hate remain invariant w.r.t. the hate target; what changes with the target is the intensity of these properties. For instance, a person may be more hurtful towards someone's race and less intense towards religion. Hence, segregating the platform-dependent component (target) from text representations can aid in learning generalizable representations.

Based on the aforementioned observations, we propose a novel \textbf{CA}usality-aware disen\textbf{T}anglement framework for \textbf{C}ross-platform \textbf{H}ate speech detection, namely, \textbf{CATCH}\footnote{\url{https://github.com/paras2612/CATCH}} 
that leverages the causal graph shown in Figure~\ref{causal_graph} to disentangle the input representations into two parts: (i) a target representation (indicating platform-dependent features) and (ii) causal representations (indicating the platform invariant features), to facilitate learning generalizable hate speech representations across different platforms. We summarize our main contributions as follows:
\begin{tcolorbox}[
  title=\textbf{ Our Contributions},
  colback=blue!5,
  colframe=blue!75!black,
  colbacktitle=blue!85!white,
  coltitle=white,
  enhanced,
  sharp corners,
  boxrule=1pt,
  fonttitle=\bfseries\large,
  ]
\begin{itemize}[itemsep = 1.0em]
  \item We propose a causal graph showing the data-generating mechanism for the hate speech detection problem.
  
  \item We propose a novel causality-aware disentanglement framework, \textbf{CATCH}, that follows the causal graph and disentangles the overall representations of the content into two parts, one for platform-dependent and one for platform invariant features, to learn generalizable representations across different platforms.
  
  \item Experimental results on four platforms demonstrate that \textbf{CATCH} outperforms the state-of-the-art baselines.
\end{itemize}
\end{tcolorbox}

\section{Related Work}
Several studies have explored generalizable hate speech detection from various angles. Some have proposed new techniques to enhance generalization~\cite{del2023socialhaterbert,kim2022generalizable,yin2022annobert}, while others have conducted thorough analyses to determine factors such as model choice, intra-dataset performance, and classification type that impact generalization~\cite{bourgeade2023did,fortuna2021well,mansur2023twitter,carvalho2023expression,antypas2023robust}. In this section, we cover methods and frameworks related to these focuses.

\subsection{Empirical Analysis of Hate Speech Generalization}
The authors of~\cite{fortuna2021well} trained four models on nine English datasets and analyzed their generalization capabilities. They concluded that the ease of generalizing varies by model and category, with labels such as ‘toxic’ or ‘offensive’ being more manageable than 'hate speech.' They also showed that the target of hate speech can influence generalization. Similarly, the authors of~\cite{bourgeade2023did} revealed how generalizability degrades across different hate speech topics, demonstrating that current state-of-the-art (SOTA) models trained on specific topics struggle to generalize to unseen ones.

Some studies specifically analyze platform-based hate speech, such as on Twitter~\cite{mansur2023twitter} and YouTube~\cite{carvalho2023expression}. The authors of~\cite{mansur2023twitter} assess various methods on Twitter, revealing how hate speech is distributed among different categories. The authors of~\cite{carvalho2023expression} examine hate against Afro-descendant, Roma, and LGBTQ+ communities within YouTube comments. Another study~\cite{antypas2023robust} emphasizes how biases in the data creation process affect model generalization. Drawing from these insights, we conclude that hate target distribution varies across platforms. Therefore, by isolating the target-dependent aspect of input representation, we can use the remaining component to create generalizable hate representations.

\subsection{Methods to Enhance Hate Speech Generalization}
Two primary techniques can be used to generalize hate speech detection in general. First, some models make use of extra or auxiliary data, such as user attributes~\cite{del2023socialhaterbert}, dataset annotator features~\cite{yin2022annobert}, or implications of hateful posts~\cite{kim2022generalizable}. By exploiting the implications of hateful posts to train contrastive pairs for a universal representation of hate content, one study~\cite{kim2022generalizable}, for instance, proposed a model that could generally recognize implicit hate speech. Another study~\cite{yin2022annobert} made the case that it is challenging for annotators to agree on subjective tasks like the identification of hate speech, and it thus suggested using the annotator's traits and the ground truth label during training to improve hate speech detection. In contrast, another study~\cite{del2023socialhaterbert} trained a model for predicting user satisfaction using data from their social context and their profiles. However, the issue with these models is that the auxiliary information they rely on might not always be easily obtainable, especially in cross-platform scenarios.

Second, some techniques use language models like BERT, which have generalization abilities because they were trained on vast text corpora. By adjusting these models using particular hate speech datasets, they can be further enhanced~\cite{caselli2020hatebert,mathew2021hatexplain}. An illustration of this is HateBERT~\cite{caselli2020hatebert}, a cutting-edge model for hate speech identification developed by optimizing a BERT model on around 1.6 million hostile messages from Reddit. There are also models like HateXplain~\cite{mathew2021hatexplain}, developed to identify explicable hate speech. In addition to these strategies, other techniques take lexical signals into account~\cite{schmidt2017survey}, such as the language employed, emotion words, POS tags in the material~\cite{markov2021exploring}, or keyphrases that target hatred~\cite{elsherief2018hate}. Another work~\cite{sheth2023peace} manually identified two causal cues, aggression and sentiment, to generalize the language model representations. 

Although these techniques have improved the ability to identify hate speech, they have their share of challenges. For instance, finetuning language models may not be feasible as they require large labeled datasets specific to the task. 
Additionally, since many social media posts contain grammatical errors like misspelled words, depending on lexical features may be ineffective. Furthermore, manually identifying cues is an exhaustive task and might overlook some important aspects, such as cues like sentiment and aggression are applicable for more explicit settings compared to implicit scenarios. To address these challenges and improve the generalization capabilities, we aim to causally disentangle the representations to segregate platform-dependent features from the overall representation, leading the remainder to be invariant and easily generalizable.

\begin{figure*}
\includegraphics[width=0.70\textwidth]{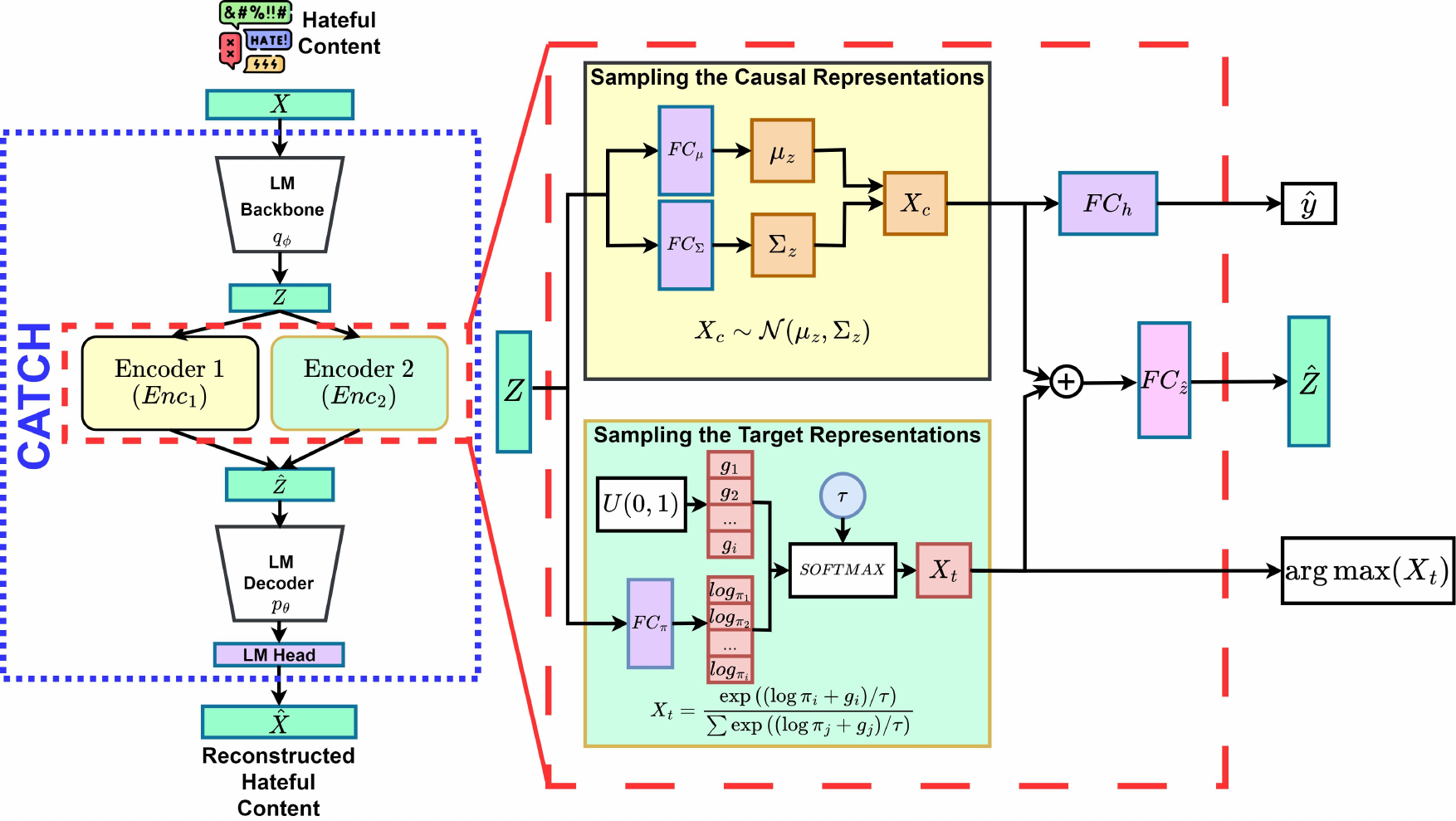}
     \caption{Overall architecture of \textbf{CATCH}. The hateful content $X$ first passes through the LM Backbone to obtain the initial input representation $Z$. The representation $Z$ then passes through two components - namely the continuous encoder $(Enc_{1})$ and the discrete encoder $(Enc_{2})$, where the continuous causal representations $X_{c}$ are generated from $\mathcal{N} (\mu_{z},\Sigma_{z})$, and the discrete target representations $X_t$ are generated from $\mathcal{F}(\pi,g)$. $X_{c}$ is used to predict the hate label $\hat{y}$, and $\arg\max(X_{t})$ represents the target label. $X_c$ and $X_t$ are concatenated to obtain the reconstructed embedding $\hat{Z}$ passed through the LM decoder to obtain the reconstructed input $\hat{X}$. $FC$ represents a fully connected layer, $U(0,1)$ represents a uniform distribution and $\tau$ represents the temperature.}
\label{fig:architecture}
\end{figure*}
\section{Proposed Method}
\subsection{Preliminaries}
Let $D_{source}$ denote a source corpus composed of a series of textual inputs (tweets or posts) $X =$ \{$x_1$, $x_2$, ..., $x_n$\}, a corresponding series of labels indicating whether the posts are hateful or not $Y =$ \{$y_1$, $y_2$, ..., $y_n$\}, and a series of labels denoting the targeted group of the post $T =$ \{$t_1$, $t_2$, ..., $t_n$\}. In the context of generalizable hate speech detection, our primary aim is to learn a mapping function $f: X \rightarrow Y$ capable of determining a generalizable representation from $D_{source}$. Such a representation enables the accurate prediction of the hate label across previously unseen target domains $D_{target}$.

Our method seeks to augment this generalization capability by adopting the disentanglement principle. This approach is inspired by recent research findings demonstrating a significant variation in the hate target across different platforms. Therefore, we opt to represent the hate speech problem through a causal graph illustrated in Figure~\ref{causal_graph}. The proposed model, \textbf{CATCH}, comprises a VAE leveraged for disentanglement and a prediction module where the disentangled representations are leveraged for prediction. The cornerstone of our method lies in disentangling the input representations into two distinct components: (1) a component ($X_{t}$) that embodies information pertinent to the hate target, and (2) a component ($X_{c}$) that encapsulates the causal aspects crucial to determine whether the content is hateful or not. This disentanglement paves the way for leveraging the generalized hate representations learned from one platform (or domain), enabling us to apply these insights to previously unseen domains. 

\subsection{Disentangling Causal and Target Representations} 
Expressing hate involves two components: the target of hate and the content that renders a post hateful. Our research proposes a causal graph and hypothesis suggesting that hate's target and intensity vary across platforms. On the other hand, the fundamental properties of hate, both abstract and quantifiable, remain constant and can be considered causal factors. Our proposed \textbf{CATCH} employs the Variational AutoEncoder (VAE) framework to capture these causal representations. The model consists of an encoder, a disentanglement module, and a decoder.

The encoder $q_{\phi}$ represents a language model, such as Roberta, which takes an input text $x$ and generates the embedding $z$ as follows:
\begin{equation}
z = q_{\phi}\left(\gamma(x)\right),
\end{equation}
where, $z \in \mathbb{R}^{S \times h_{d}}$ represents the language model representation, $\gamma(x)$ represents the tokenizing function of the language model, $S$ indicates the sequence length, and $h_{d}$ is the embedding dimension. We utilize the embedding of the start-of-sentence token ([CLS]) located at position $0$, denoted as $z^{[CLS]} \in \mathbb{R}^{h_{d}}$, as the input representation for $x$.

\textbf{Disentangling the Causal Component}\quad To disentangle $z$ and obtain the causal counterpart $X_{c}$, we employ a VAE architecture. Disentangled representation refers to a factorized representation, where each latent variable corresponds to a single explanatory variable responsible for generating the data. VAEs are well-suited for achieving this task, as they impose a standard Gaussian prior distribution on the latent space and approximate the posterior using a parameterized neural network. Two feedforward neural networks, $FC_{\mu}$ and $FC_{\Sigma}$, are employed to map $z$ to the Gaussian distribution parameters of hate. The latent representation $X_{c} \in \mathbb{R}^{h_{causal}}$ is sampled from the Gaussian distribution defined by the corresponding $(\mu_{z}, \Sigma_{z})$ using the re-parameterization trick~\cite{kingma2014semi}, where $h_{causal}$ represents the hidden dimension size:
\begin{equation}
X_{c}= Enc_{1}(\mu_z,\Sigma_z) = \mu_{z} + \Sigma_{z} \odot \epsilon \sim \mathcal{N}(\mathbf{0}, \mathbf{I}).
\end{equation}
Here, $\mu_{z} = FC_{\mu}(z)$, $\Sigma_{z} = FC_{\Sigma}(z)$, and $\sim \mathcal{N}(\mathbf{0}, \mathbf{I})$ represents the normal distribution, and $\mathbf{I}$ represents unit variance.

\textbf{Disentangling the Target component}\quad Besides the causal component, $z$ also includes platform-dependent features, among which is the target of hate. Recent studies reveal variations in hate speech across different platforms. For instance, YouTube comments predominantly manifest hate based on sexism~\cite{doring2019male}, while Twitter is more affected by racism-based hate~\cite{waseem2016you}.

However, unlike the causally invariant hate features, the platform-dependent features related to the target are discrete. Hate speech targets can be categorized into eight broad classes: Race, Religion, Gender, Sexual Preference, Nationality, Immigration, Disability, and Class~\cite{bourgeade2023did}. Consequently, the latent representation for the target is less likely to be Gaussian and more likely to be categorical or discrete in nature~\cite{feng2021shot}. To effectively model the discrete latent variable, we employ VAEs leveraging the "Gumbel-Softmax" trick~\cite{jang2016categorical}. This enables the reparameterization of discrete variables within the VAE framework, facilitating the use of backpropagation for optimization and learning a disentangled representation~\cite{jang2016categorical}.

The process commences by using a feed-forward neural network $FC_{\pi}$ to map $z$ to the latent representation $z_{\pi} \in \mathbb{R}^{h_{disc}}$ for the discrete variable, where $h_{disc}$ represents the hidden dimensions. $z_{\pi}$ is obtained as $z_{\pi}=FC_{\pi}(z)$.

Let $z_{\pi}$ represent the categorical variable with class probabilities \{$\pi_{1}, \pi_{2}, \cdots \pi_{h_{disc}}$\}. We then apply the logarithmic function to each probability and add Gumbel noise, sampled by taking two logs of a uniform distribution. Let $g_{i}$ represent the i.i.d. samples drawn from Gumbel (0,1)~\cite{jang2016categorical}. 
To ensure a continuous, differentiable approximation 
, and generate $h_{disc}$-dimensional sample vectors, we further leverage the Softmax function to obtain $X_{t}$:
\begin{equation}
X_{t} = Enc_{2}(\pi,g) =  \frac{\exp\left(\left(\log\left(\pi_i\right) + g_i\right) / \tau\right)}{\sum_{j=1}^{h_{disc}} \exp\left(\left(\log\left(\pi_j\right) + g_j\right) / \tau\right)} \quad \text{for } i = 1, \ldots, h_{disc}.
\end{equation}
where $\tau$ represents the temperature.

\textbf{Reconstructing the Input from the Disentangled components}\quad
To facilitate the training, we aim to reconstruct the input from the obtained components $X_{c}$ and $X_{t}$. To do so, we first concatenate them together as $[X_{c}$ $\mid$ $X_{t}]$ where $[\mid]$ is the concatenation operation. Then we pass it through another feed-forward neural network, namely, $FC_{\hat{z}}$. The obtained representation after concatenation is given by $\hat{z} = FC_{\hat{z}}([X_{c}||X_{t}])$. To recreate the input, we feed $\hat{z}$ through a Language Model (LM) decoder $p(x|\hat{z})$.  We utilize BART-base decoder~\cite{wolf2019huggingface} as the LM-decoder, as it shares the vocabulary with RoBERTa~\cite{wolf2019huggingface} implementations and is proved powerful in many generative tasks. The obtained tokens are given as follows:
\begin{equation}
    \hat{x} = LM Head(p_{\theta}(\hat{z}))
\end{equation}
where $LM Head$ is a feed-forward layer to map the decoder ($p_{\theta})$ embeddings into tokens. The reconstructed loss is computed between the input token ids and the reconstructed token ids and is formulated as follows:
\begin{equation}
    \mathcal{L}_{recon}(\gamma(x), \hat{x}) = - \sum_{i=1}^{S} \gamma(x) \log(\hat{x}_{i})
\end{equation}
where $\mathcal{L}$ represents the cross entropy loss, and $S$ is the sequence length. To ensure that both the disentangled latent spaces' posteriors are close to their prior distribution, we further enforce KL-divergence losses for both latent spaces. The Evidence Lower BOund (ELBO) for the disentanglement module is formulated as follows:
\begin{equation}
    \mathcal{L}_{VAE} = \mathcal{L}_{recon} + \alpha_{t}*\operatorname{L}_{\mathbb{D}_{target}} + \alpha_{c}*\operatorname{L}_{\mathbb{D}_{causal}},
\end{equation}
where $\alpha_{t}$ represents the coefficient that controls the contribution of the KL loss for the target, and $\alpha_{c}$ represents the coefficient that controls the contribution of the causal KL loss. To facilitate learning from the target labels, we follow~\cite{kingma2014semi} and add a cross-entropy loss to the target's KL loss. $\operatorname{KL}_{\mathbb{D}_{target}}$ is given by,
\begin{equation}
\begin{aligned}
\operatorname{L}_{\mathbb{D}_{target}} &= -D_{\mathrm{KL}}\left(Enc_{2}(X_{t} \mid X) \| p(X_{t})\right) + \alpha_{tc}*\mathcal{L}_{CE}(\arg\max(X_{t}),t)
\end{aligned}
\end{equation}
where $t$ is the ground truth target label,  $\arg \max(X_{t})$ is the predicted target label given by, $D_{\mathrm{KL}}$ represents the KL divergence, $\mathcal{L}_{CE}$ is the cross entropy loss and $\alpha_{tc}$ is the coefficient controlling the contribution of the target classification loss. Similarly, $\operatorname{KL}_{\mathbb{D}_{target}}$ is given by, 
\begin{equation}
\begin{aligned}
\operatorname{L}_{\mathbb{D}_{causal}} &= -D_{\mathrm{KL}}\left(Enc_{1}(X_{c} \mid X) \| p(X_{c})\right)
\end{aligned}
\end{equation}
\subsection{Model Training}
For hate speech detection, the disentangled latent causal representation $X_c$ is utilised to compute the classification probability:
\begin{equation}
\hat{y}_i=\operatorname{Softmax}(FC_{h}(X_{c}))    
\end{equation}
where $FC_{h}$ represents a fully connected layer for hate classification. Then we compute the overall loss $\mathcal{L}$ using standard cross-entropy:
\begin{equation}
\mathcal{L}_{hate}=-\frac{1}{N} \sum_{i=1}^{|D_{source}|} y_i \log \hat{y}_i    
\end{equation}
where $y_i$ represents the true hate label, and $\hat{y}_i$ denotes the predicted hate labels. Finally, we combine all proposed modules and train in a multi-task learning manner:
\begin{equation}
\mathcal{L}=\mathcal{L}_{hate}+\mu_d \mathcal{L}_{VAE}
\end{equation}
where the $\mu_d$ is a pre-defined weight coefficient.
\begin{table*}[]
\begin{tabular}{cccccccc}
\hline
\multirow{2}{*}{\textbf{Source}}  & \multirow{2}{*}{\textbf{Target}} & \multicolumn{6}{c}{\textbf{Models}}                                            \\ \cline{3-8} 
 &
   &
  \textbf{\begin{tabular}[c]{@{}c@{}}Easy\\ Mix\end{tabular}} &
  \textbf{\begin{tabular}[c]{@{}c@{}}Hate\\ Bert\end{tabular}} &
  \textbf{\begin{tabular}[c]{@{}c@{}}Hate\\ Xplain\end{tabular}} &
  \textbf{\begin{tabular}[c]{@{}c@{}}POS+\\ EMO\end{tabular}} &
  \textbf{PEACE} &
  \textbf{CATCH}  \\ \hline
\multirow{4}{*}{\textbf{GAB}}     & \textbf{GAB}                     & 0.70       & \textbf{0.89} & {\ul 0.87}    & 0.77 & 0.76       & 0.82          \\
                                  & \textbf{Reddit}                  & 0.62       & 0.66          & 0.66          & 0.56 & {\ul 0.69} & \textbf{0.72} \\
                                  & \textbf{Twitter}                 & 0.64       & 0.63          & {\ul 0.65}    & 0.44 & 0.64       & \textbf{0.69} \\
                                  & \textbf{YouTube}                 & 0.62       & 0.60          & 0.62          & 0.50 & {\ul 0.64} & \textbf{0.66} \\ \hline
\multirow{4}{*}{\textbf{Reddit}}  & \textbf{GAB}                     & 0.51       & 0.52          & {\ul 0.56}    & 0.45 & 0.55       & \textbf{0.58} \\
                                  & \textbf{Reddit}                  & {\ul 0.95} & \textbf{0.98} & 0.94          & 0.91 & 0.90       & 0.86          \\
                                  & \textbf{Twitter}                 & 0.54       & 0.51          & 0.54          & 0.43 & {\ul 0.55} & \textbf{0.60} \\
                                  & \textbf{YouTube}                 & 0.64       & 0.69          & 0.60          & 0.57 & {\ul 0.70} & \textbf{0.76} \\ \hline
\multirow{4}{*}{\textbf{Twitter}} & \textbf{GAB}                     & 0.62       & 0.63          & 0.62          & 0.56 & {\ul 0.65} & \textbf{0.67} \\
                                  & \textbf{Reddit}                  & 0.64       & 0.62          & 0.62          & 0.48 & {\ul 0.66} & \textbf{0.69} \\
                                  & \textbf{Twitter}                 & 0.67       & \textbf{0.86} & {\ul 0.83}    & 0.68 & 0.63       & 0.78          \\
                                  & \textbf{YouTube}                 & {\ul 0.65} & 0.59          & 0.63          & 0.53 & 0.64       & \textbf{0.68} \\ \hline
\multirow{4}{*}{\textbf{YouTube}} & \textbf{GAB}                     & 0.44       & \textbf{0.62} & 0.47          & 0.43 & 0.48       & {\ul 0.56}    \\
                                  & \textbf{Reddit}                  & 0.67       & 0.65          & 0.62          & 0.56 & {\ul 0.69} & \textbf{0.72} \\
                                  & \textbf{Twitter}                 & 0.45       & {\ul 0.59}    & 0.56          & 0.49 & 0.58       & \textbf{0.64} \\
                                  & \textbf{YouTube}                 & {\ul 0.86} & 0.84          & \textbf{0.88} & 0.64 & {\ul 0.86} & 0.79          \\ \hline
\end{tabular}
\caption{Cross-platform and in-dataset evaluation results for the different baseline models compared against CATCH. Boldfaced values denote the best performance, and the underline denotes the second-best performance.}
\label{tab:results}
\end{table*}

\section{Experiments}
In this section, we implement a set of experiments designed to validate the effectiveness of \textbf{CATCH} for learning generalizable representations for detecting hate speech with causality-aware disentanglement. We make use of multiple datasets procured from various online platforms to provide a comprehensive evaluation of our methodology. Further, we conduct an extensive analysis encompassing ablation tests, and interpretability analysis to dissect the inner workings of our model. Moreover, we conduct a case study comparing how \textbf{CATCH} performs in comparison with the large scale language models such as Falcon~\cite{refinedweb}, and GPT-4~\cite{peng2023instruction}. The objective of these experimental analyses is to answer the following set of pertinent research questions:
\begin{itemize}
    \item \textbf{RQ.1} Can the disentanglement of the input representation into a causal component and a platform-dependent component (target) aid in learning invariant causal representations that can improve the generalizability of hate speech detection?
    \item \textbf{RQ.2} Are the learned disentangled representations invariant across the different platforms?
    \item \textbf{RQ.3} What is the contribution of \textbf{CATCH's} components in aiding with the generalizability of representations?    
\end{itemize}

\begin{table}[H]
\centering
\footnotesize
\begin{tabular}{cccc}
\toprule
\textbf{Datasets} & \textbf{No. of Posts} & \textbf{Hateful Posts} & \textbf{Hate \%} \\ 
\midrule
\textbf{GAB}~\cite{mathew2021hatexplain}      & 11,093                & 8,379                  & 75.5             \\ 
\textbf{Reddit}~\cite{kennedy2020constructing}   & 37,164                & 10,562                 & 28.4             \\ 
\textbf{Twitter}~\cite{mathew2021hatexplain}  & 9,055                 & 2,406                  & 26.5             \\ 
\textbf{YouTube}~\cite{salminen2018anatomy}  & 1,026                 & 642                    & 62.5             \\ \bottomrule
\end{tabular}
\vspace{1mm}
\caption{Dataset statistics with corresponding platforms and percentage of hateful comments or posts.}
\label{datasets}
\end{table}

\subsection{Datasets and Evaluation Metrics}
We perform binary classification of hate speech detection on widely used benchmark hate datasets. Since we aim to verify cross-platform generalization, we use data from four different platforms for cross-platform evaluation: GAB, Reddit, Twitter, and YouTube. All datasets are in the English language. GAB~\cite{mathew2021hatexplain} is a collection of annotated posts from the GAB website. It consists of binary labels indicating whether a post is hateful or not. These instances are annotated with corresponding explanations, where crowd workers justify why the given post, or content is considered hateful. 

Reddit~\cite{kennedy2020constructing} is a collection of posts indicating whether it is hateful or not. It contains ten ordinal labels (sentiment, (dis)respect, insult, humiliation, inferior status, violence, dehumanization, genocide, attack/defense, hate speech), which are debiased and aggregated into a continuous hate speech severity score (hate speech score). We binarize this data so that any data with a hate speech score less than 0.5 is considered non-hateful. Twitter~\cite{mathew2021hatexplain} contains instances of hate speech gathered from tweets on the Twitter platform. Similar to the Gab dataset, these instances are also paired with explanations written by crowd workers, aiming to explain the hatefulness present in the respective tweets. Finally, YouTube~\cite{salminen2018anatomy} is a collection of hateful expressions and comments posted on the YouTube platform. All these datasets contain the hate labels as well as the target labels. A summary of the datasets can be found in Table \ref{datasets}. We use macro F1-measure for validation.

\subsection{Experimental Setting}
\noindent{\textbf{Implementation Details}}\quad
We trained our framework \textbf{CATCH} using RoBERTa-base for the LM-encoder and BART-base for the LM-decoder with the Huggingface Transformers library. We optimized the model with cross-entropy loss and AdamW, using a learning rate of 0.0001, dropout rate of 0.2, and parameters $\alpha_{t}$ as 0.05, $\alpha_{c}$ as 0.05, $\alpha_{tc}$ as 0.001, and $\mu_{d}$ as 0.5. Training was conducted on an NVIDIA GeForce RTX 3090 GPU with 40 GB VRAM, with early-stopping.

\noindent{\textbf{Baselines}}\quad We compare our \textbf{CATCH} framework against various state of the art baselines. These baseline methods were designed to enhance the generalization prowess for cross-platform hate speech detection. The details of the methods categories is shown below:
\begin{itemize}
    \item \textbf{EasyMix}~\cite{roychowdhury2023data} leverages data augmentation to generate training samples that cover a wide distribution of hateful language.
    \item The \textbf{POS+EMO} baseline~\cite{markov2021exploring} uses linguistic cues like POS tags, stylometric features, and emotional cues.
    \item \textbf{HateBERT}~\cite{caselli2020hatebert} uses over 1.5 million Reddit messages from suspended communities known for encouraging hate speech to fine-tune the BERT-base model. We further finetune HateBERT on each source domain and report the performance.
    \item Utilizing hate speech detection datasets from Twitter and Gab, \textbf{HateXplain}~\cite{mathew2021hatexplain} is improved with an emphasis on a three-class classification task (hate, offensive, or normal) with human-annotated justifications. We also fine-tune HateXplain on each source domain and then report its performance.
    \item \textbf{PEACE}~\cite{sheth2023peace} leverages two causal cues, namely, the sentiment and aggression, to make representations more generalizable. 
\end{itemize}

\subsection{Performance Comparisons (RQ.1)}
In our study, using a variety of real-world datasets, we compare various baseline models with \textbf{CATCH} on four different platforms. Each dataset is divided into train and test sets to evaluate the generalizability of these models. Then, each model is trained using data obtained from a single platform and assessed against test sets obtained from all platforms. The performance of several test sets is compared using the macro-F1 measure as a benchmark in Table \ref{tab:results}. The \textbf{Target Platforms} column lists the platforms used for model assessment, while the \textbf{Source Platform} column provides the platform used for model training. We note the performance of each model for each source dataset in cross-platform and within-platform situations. 
We discern the following observations regarding the cross-platform performance regards to RQ.1:
\begin{itemize}
    \item Predominantly, the \textbf{CATCH} delivers the most efficient performance in a cross-platform scenario. On average, the Model enhances cross-domain efficiency by approximately 3\% on GAB, 5\% on Reddit, 3\% on Twitter, and 3\% on YouTube. We attribute this improvement to applying causality-aware disentanglement, which initially distinguishes the platform-dependent target from the platform-invariant causal representations, utilizing only the causal representations for hate prediction. Moreover, these results further corroborate the hypothesis that the intensity of hate varies with target, while the essential hate attributes remain consistent.
    \item PEACE outperforms other models, including HateBERT, primarily due to its use of invariant causal indicators like sentiment and aggression for representation, which fosters better generalization. Additionally, fine-tuning BERT models, as seen with HateBERT and HateXplain, on hate speech datasets improves performance. However, PEACE is not entirely superior because not all hate speech includes sentiment and aggression~/cite{perez2017racism}, and it only considers two attributes, omitting others like context and socio-political factors that are also crucial~/cite{garg2022handling}.
    \item The EasyMix baseline encounters challenges in generalizing in certain cases. It employs data augmentation and frames the hate speech task as an entailment issue. Nevertheless, we speculate that its difficulty in outperforming other models arises from its reliance on varied combinations of the samples in the source domain, such as mixing non-hateful samples with other non-hateful samples. Such an approach might enhance the dependence on incidental correlations, thus benefiting in-domain performance but negatively impacting out-of-domain performances.
    \item The baseline based on linguistic features (POS + EMO) doesn’t generalize well to these datasets. We postulate that this is due to the highly unstructured and grammatically incorrect nature of the posts in these datasets. Even following pre-processing, the inferred POS tags and emotion words may not accurately reflect the hateful content. Consequently, the reliance on these features impairs the generalization performance.
\end{itemize}
\begin{figure}[h]%
    \centering
    \subfloat[\centering \textbf{CATCH}] 
    {{\includegraphics[width=0.19\textwidth]{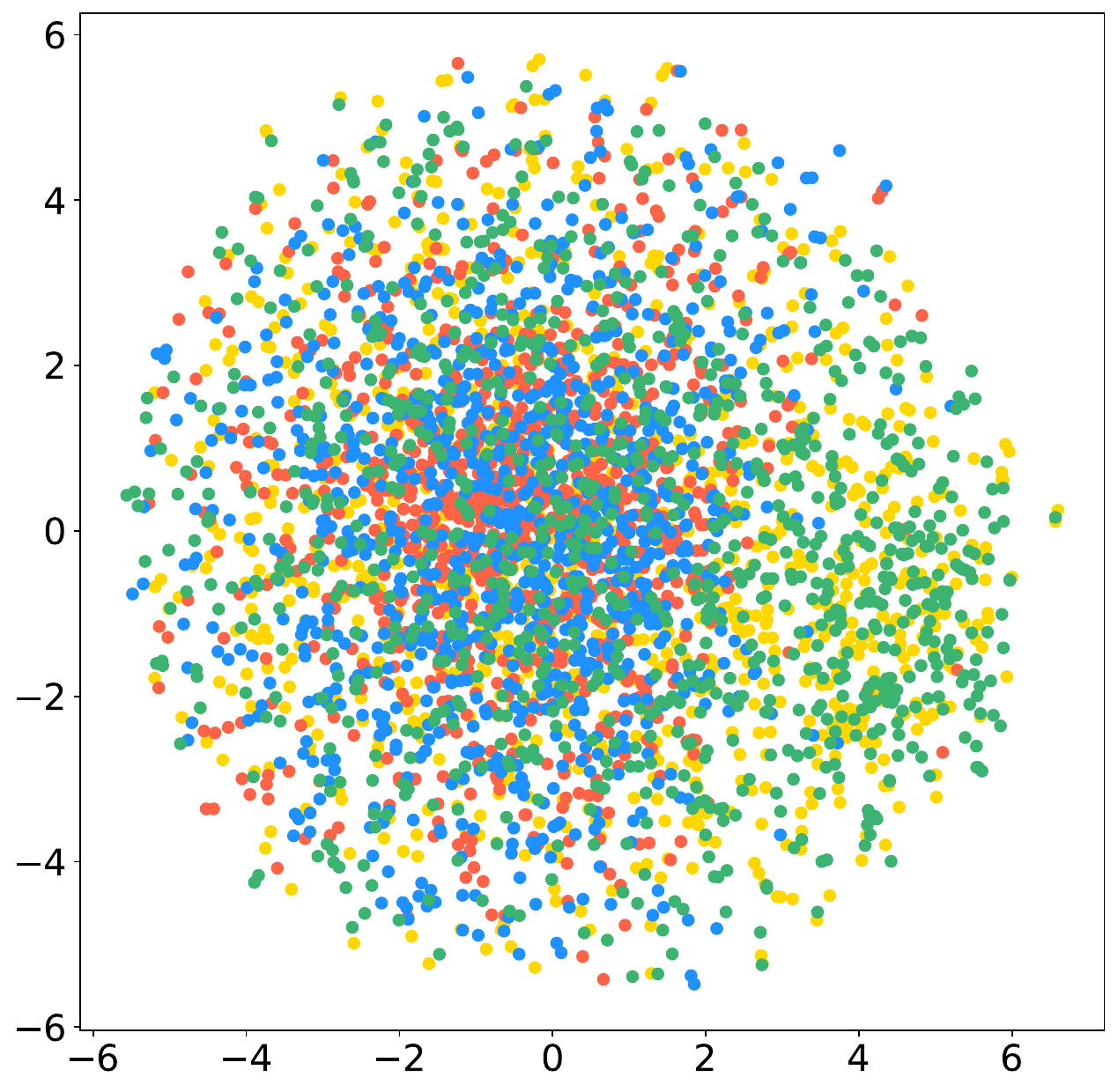} 
    \label{gab_model}}%
     }%
    \subfloat[\centering HateBERT]
     {{\includegraphics[width=0.19\textwidth]{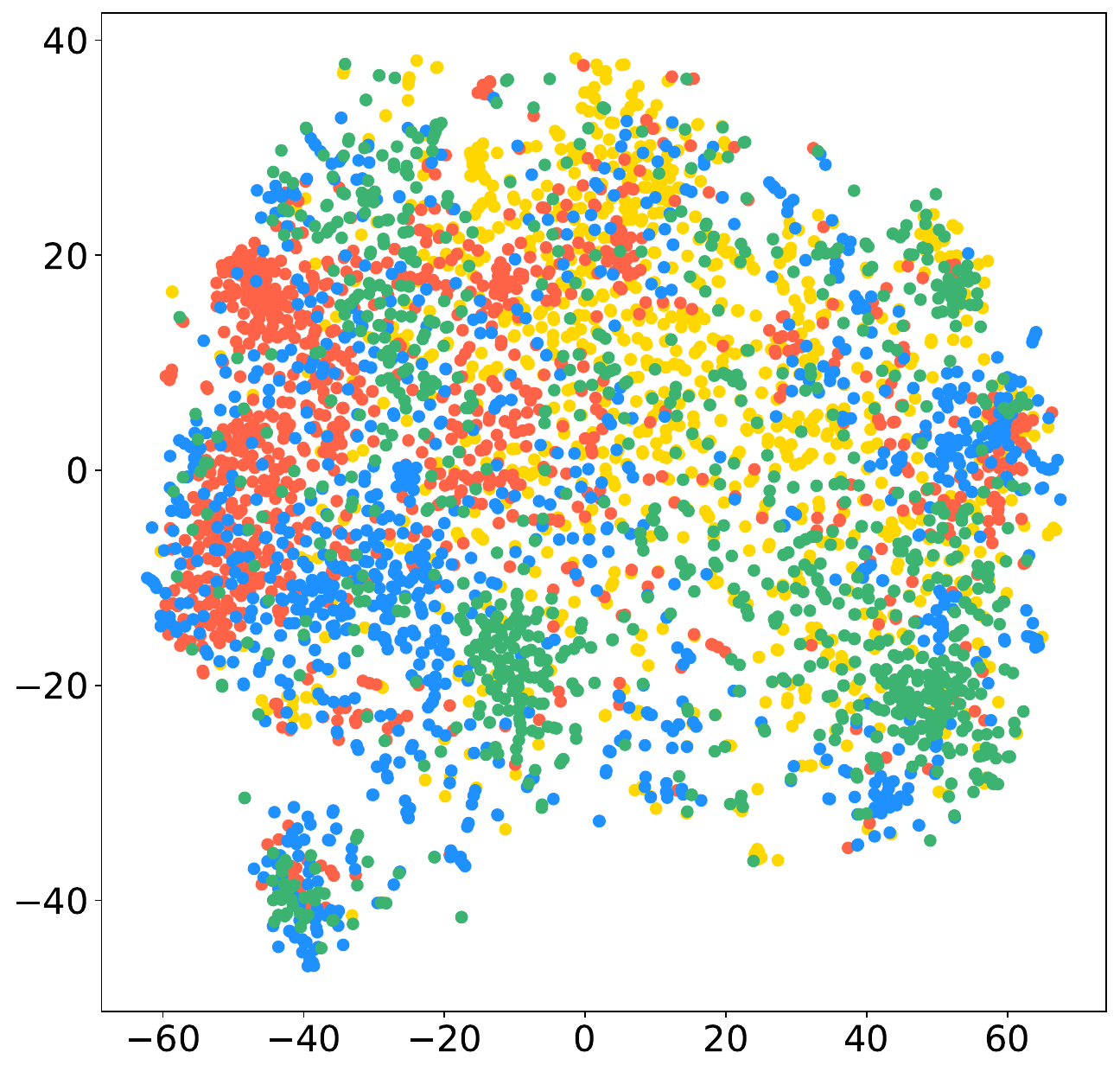} 
    \label{gab_HB}}%
    }\\%
    \subfloat[\centering PEACE]
     {{\includegraphics[width=0.19\textwidth]{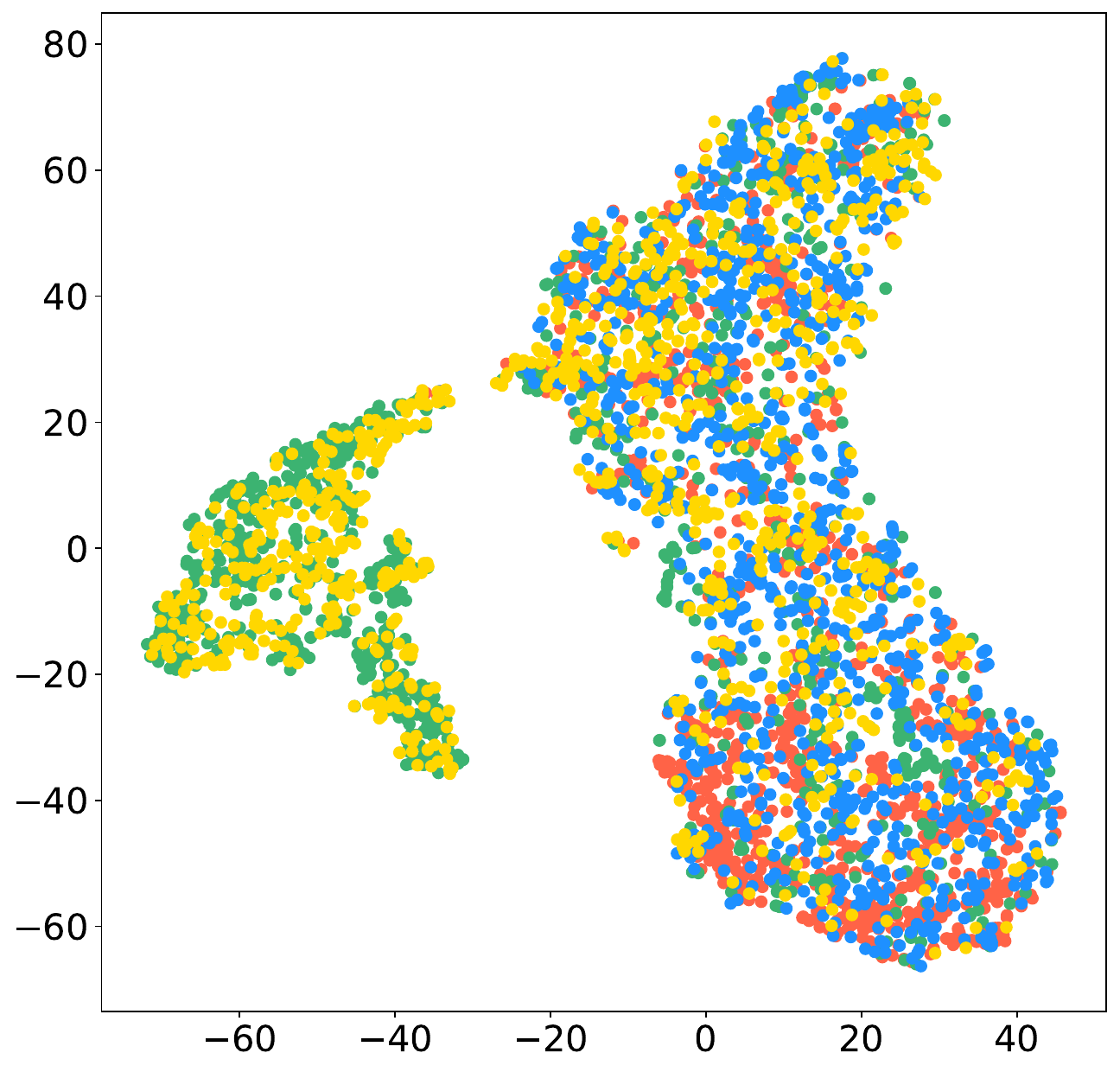} 
    \label{gab_PEACE}}%
    }
    \subfloat[\centering HateXplain]
     {{\includegraphics[width=0.19\textwidth]{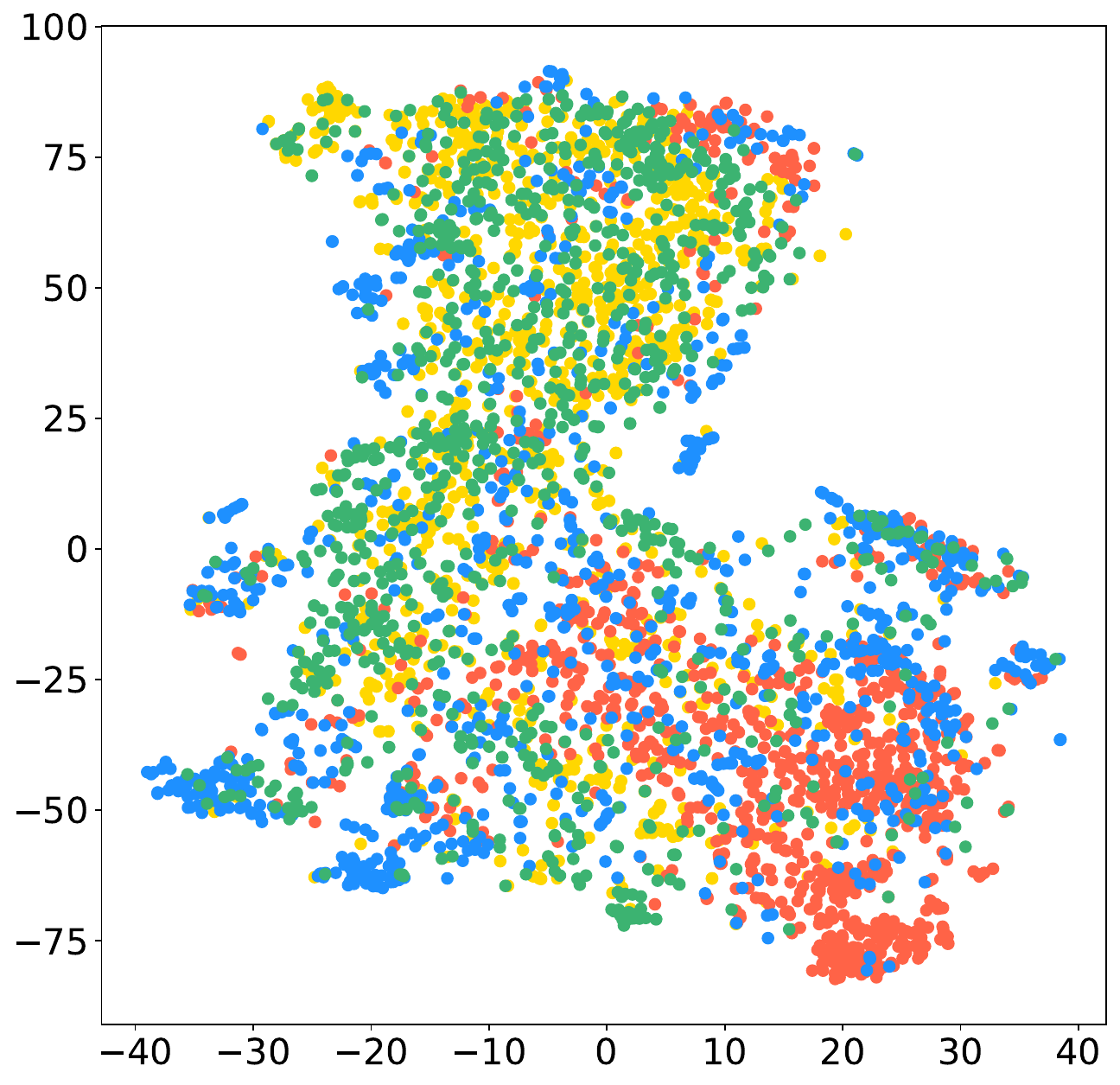} 
    \label{gab_HE}}%
    }
    
    \subfloat
     {{\includegraphics[width=\columnwidth]{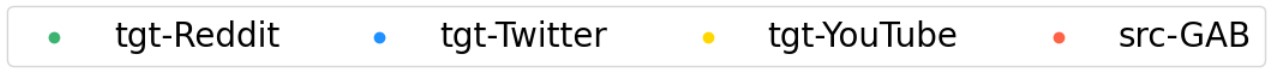} 
    \label{legend}}%
    }
    \caption{Measuring whether the disentangled causal representations are invariant across the different platforms. "src" denotes the source and "tgt" denotes the target platforms.}
    \label{invar}
\end{figure}
\subsection{Are the Disentangled Causal Representations Invariant? (RQ.2)}
One of the fundamental characteristics of the \textbf{CATCH} is its capacity to disentangle input representations into causal and platform-dependent (target) elements. The causal representations are expected to exhibit invariance as they are employed to predict the hate label. To validate the presence of this invariance, we carry out an additional experiment. In this, we train the \textbf{CATCH} on GAB, which serves as our source platform. We then examine the t-SNE plot of the causal representations acquired from various platforms by randomly sampling 1,000 instances of hateful and non-hateful posts. The extent of overlap between these representations indicates the model's ability to extract invariant features, which are crucial for generalization~\cite{mahajan2021domain}. For an equitable comparison, we extend this experiment to HateBERT, PEACE, and HateXplain, our top three baseline models, and the results are illustrated in Figure~\ref{invar}.

As can be discerned from the figure, the causal representations generated by the \textbf{CATCH} display a high degree of overlap compared to those produced by HateBERT, HateXplain, and PEACE. This suggests that the \textbf{CATCH} is adept at learning invariant features universally shared across the domains. Furthermore, PEACE's representations exhibit more overlap than those of HateBERT and HateXplain. This is consistent with the fact that PEACE identifies two intrinsic causal cues common to various platforms. Despite this, PEACE does not outperform the \textbf{CATCH}. This could be because while PEACE manually identifies two cues, there might be other unidentified cues (e.g., context) that could provide additional insights into whether the content is hateful. Although both HateBERT and HateXplain have some overlap, there is very little overlap across some platform combinations. For instance, both these models have little overlap in terms of representations for GAB and YouTube, whereas both Reddit and Twitter have high overlap.

\subsection{Contribution of the Components (RQ.3)}
The \textbf{CATCH} architecture is a complex construct composed of various integrated modules and losses. To assess the influence of these individual elements on cumulative performance, we have conducted an experiment featuring four distinct variations of the \textbf{CATCH}. These variations include: (i) \textit{CATCH w/o Hate \& Target Loss}, which excludes hate and target losses, (ii) \textit{CATCH w/o finetuning}, which omits the fine-tuning for the LM encoder and decoder, (iii) \textit{CATCH w/o Hate Loss}, which forgoes the Hate Loss, and (iv) \textit{CATCH w/o Target Loss}, which disregards the Target Loss.

\begin{figure}[h]%
    \centering
    \vspace{-7mm}
    \subfloat[\centering Reddit] 
    {{\includegraphics[width=0.38\textwidth]{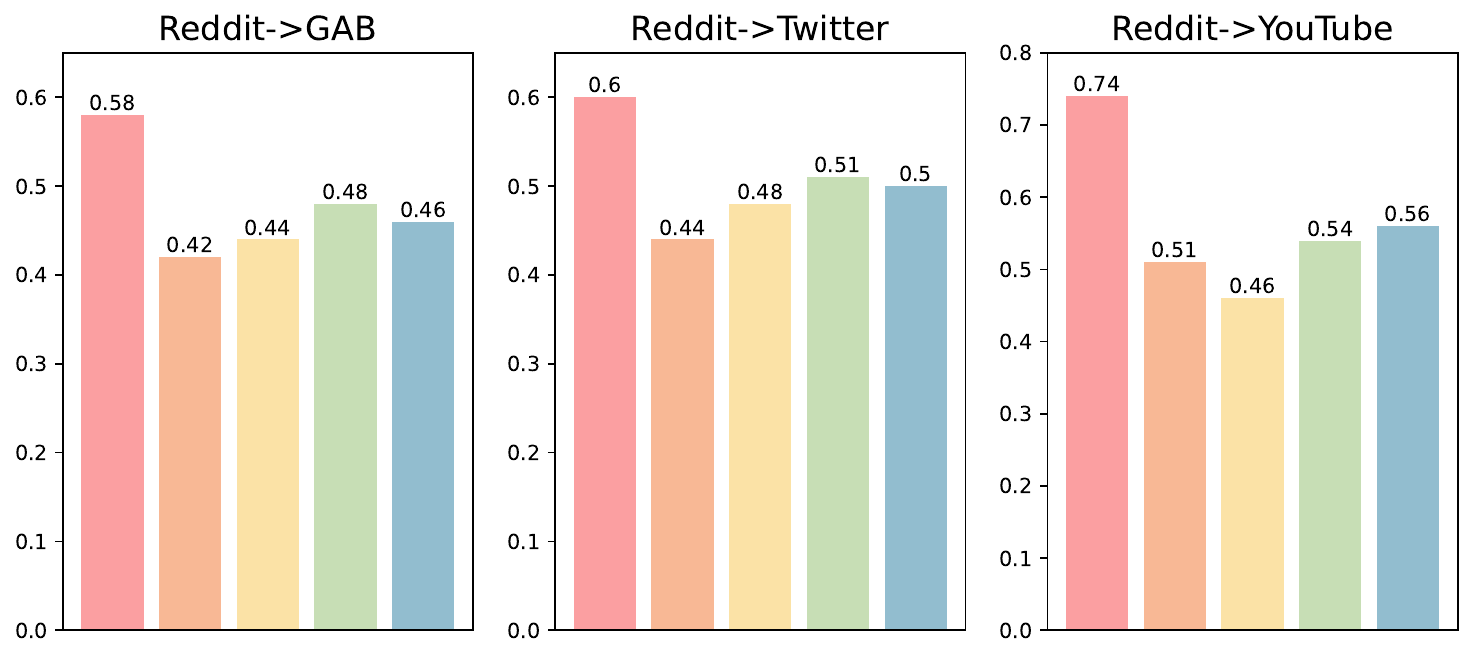} 
    \label{ablationReddit}}%
     }%
     \\ 
    \subfloat[\centering Twitter]
     {{\includegraphics[width=0.38\textwidth]{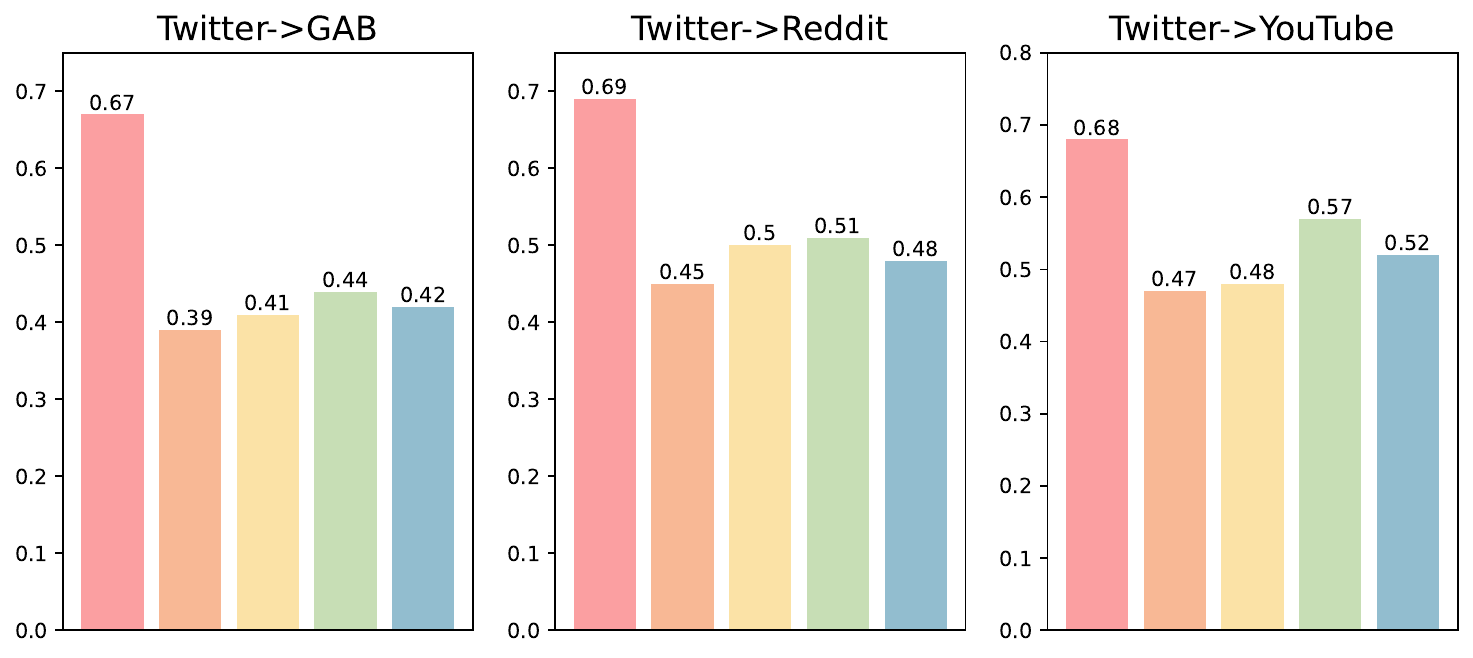} 
    \label{ablationTwitter}}%
    }%
    \\
    \subfloat
     {{\includegraphics[width=\columnwidth]{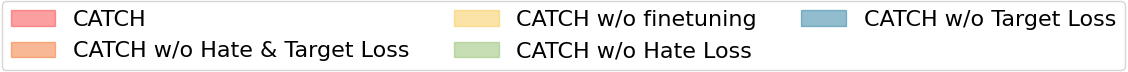} 
    \label{legend2}}%
    }
    \caption{Macro-F1 score reflecting the importance of each cue compared with the final model for Reddit and Twitter.}
    \label{ablation}
\end{figure}

We trained these variations using the Reddit and Twitter datasets in cross-platform experiments. The outcomes of these experiments are illustrated in Figure \ref{ablation}(a) for the Reddit dataset and Figure \ref{ablation}(b) for the Twitter dataset. The results demonstrate that the \textbf{CATCH} delivers optimal performance when both the fine-tuning process and the various loss components are incorporated. Neglecting these components results in a 10-18\% performance drop.

Among the variations, the most significant performance benefit for the \textbf{CATCH} comes from including target and hate losses. Incorporating target and hate losses significantly enhances the \textbf{CATCH} model's performance by guiding the disentanglement of components to capture unique, generalizable information. Performance drops notably without these losses or when the language model encoder and decoder are not fine-tuned on hateful content.

\begin{table}[H]
\footnotesize
\begin{tabular}{ccccc}
\toprule
\multirow{2}{*}{\textbf{Models}} & \multicolumn{4}{c}{\textbf{Target   Platforms}}                                         \\ \cmidrule{2-5} 
                      & \multicolumn{1}{c}{\textbf{GAB}}  & \multicolumn{1}{c}{\textbf{Reddit}} & \multicolumn{1}{c}{\textbf{Twitter}} & \textbf{YouTube} \\ 
\textbf{GPT4}         & \multicolumn{1}{c}{\textbf{0.64}} & \multicolumn{1}{c}{0.66}            & \multicolumn{1}{c}{\textbf{0.67}}    & 0.63             \\ 
\textbf{Falcon}                 & \multicolumn{1}{c}{0.42} & \multicolumn{1}{c}{0.58} & \multicolumn{1}{c}{0.54} & 0.55 \\ 
\textbf{CATCH (Avg.)} & \multicolumn{1}{c}{0.61}          & \multicolumn{1}{c}{\textbf{0.71}}   & \multicolumn{1}{c}{0.64}             & \textbf{0.70}     \\ 
\bottomrule
\end{tabular}
\vspace{1mm}
\caption{Performance comparison of LLMs, GPT4 and Falcon, with CATCH for generalizable hate speech detection.}
\label{tab:llms}
\end{table}

\section{Are LLMs Generalizable When It Comes To Hate Speech? -- A Case Study}

Recent advancements in natural language processing (NLP) have been driven by large language models~\cite{koubaa2023gpt,kocon2023chatgpt,li2023chatgpt} (LLMs) like GPT-4 and Falcon, which excel in various applications due to their high capacity and extensive data training. However, their effectiveness in complex tasks (those that rely on comprehension of complicated social settings) such as hate speech detection is not well-established. 

However, to test this hypothesis, we conducted a performance comparison of \textbf{CATCH} with the leading-edge LLMs, GPT4~\cite{OpenAI2023GPT4TR} and Falcon~\cite{refinedweb}, as shown in Table \ref{tab:llms}. We give the following prompt to GPT4 and Falcon, and note the label predictions generated by these models. We provide a few instances of hateful and non-hateful content from each platform to leverage the model's capabilities to easily do the prediction. The outcomes show how well the models generalizes across the platforms.


\begin{tcolorbox}[
  title=\textbf{Prompt to Detect Hate},
  colback=blue!10,
  colframe=blue!75!black,
  colbacktitle=blue!85!white,
  coltitle=white,
  enhanced,
  sharp corners,
  boxrule=1pt,
  fonttitle=\sffamily\bfseries\large,
  ]
\small
\textsf{Here are some examples of hateful content on \texttt{\textbf{Platform}}: \texttt{\textbf{samples}}. Based on the given examples, decide whether the following text is hateful or not? Just answer in Yes or No.}
\end{tcolorbox}

While GPT-4 showed strong results in some cases, its performance was inconsistent, particularly on Reddit and YouTube. Falcon's results were slightly inferior. Conversely, \textbf{CATCH} outperformed both on these platforms, suggesting that while LLMs have potential,\textbf{CATCH} causality-aware approach may offer a more nuanced solution for platform-specific hate speech detection.


In conclusion, although LLMs like GPT4 and Falcon have potential, they may fall short in nuanced tasks like hate speech detection. \textbf{CATCH}, however, uses causality-aware disentanglement, making it more effective in dealing with platform-specific variations in hate.

\section{Conclusion and Future Work}
Targeted hate speech, a growing issue on social media platforms, necessitates the development of effective automated detection techniques. Addressing the challenges of hate speech proliferation and scarce labeled data, our study presents the \textbf{CATCH}, a novel framework that disentangles causal (platform invariant) and target (platform-dependent) components from input texts for enhanced hate speech detection. By separating platform-dependent traits, the invariant component aids successful generalization across diverse platforms. Experiments confirmed \textbf{Model's} superiority over existing baselines and verified the invariant nature of its learned causal representations. Additional case studies revealed insights into the performance of large-scale language models like GPT4 and Falcon in generalizable hate speech detection.

However, \textbf{CATCH}'s reliance on target labels for representation segregation poses a challenge, given that such labels may not always be available. Future work could focus on enhancing the \textbf{CATCH} framework to operate effectively without target labels. The insights and results from this study lay a promising groundwork for creating a safer and more respectful digital social environment.

\section{Acknowledgements}
This material is based upon work supported by, or in part by the Office of Naval Research (ONR) under
contract/grant number N00014-21-1-4002 and National Science Foundation (NSF) (\#2311716).
\section{Ethical Statement}

\subsection{Freedom of Speech and Censorship}
 
Our research aims to develop algorithms that can effectively identify and mitigate harmful language across multiple platforms. We recognize the importance of protecting individuals from the adverse effects of hate speech and the need to balance this with upholding free speech. Content moderation is one application where our method could help censor hate speech on social media platforms such as Twitter, Facebook, Reddit, etc. However, one ethical concern is our system's false positives, i.e., if the system incorrectly flags a user's text as hate speech, it may censor legitimate free speech. Therefore, we discourage incorporating our methodology in a purely automated manner for any real-world content moderation system until and unless a human annotator works alongside the system to determine the final decision. 

\subsection{Use of Hate Speech Datasets}

In our work, we incorporated publicly available well-established datasets. We have correctly cited the corresponding dataset papers and followed the necessary steps in utilizing those datasets in our work. We understand that the hate speech examples used in the paper are potentially harmful content that could be used for malicious activities. However, our work aims to help better investigate and help mitigate the harms of online hate. Therefore, we have assessed that the benefits of using these real-world examples to explain our work better outweigh the potential risks.

\subsection{Fairness and Bias in Detection}

Our work values the principles of fairness and impartiality. To reduce biases and ethical problems, we openly disclose our methodology, results, and limitations and will continue to assess and improve our system in the future.

\bibliographystyle{ACM-Reference-Format}
\balance
\bibliography{sample-base}


\end{document}